%% file: main.tex
\RequirePackage{fix-cm}
\documentclass[smallcondensed]{svjour3}     
\smartqed  
\usepackage{amsmath} 
\usepackage{amssymb}  

\usepackage{comment}
\usepackage{graphicx}

\usepackage{adjustbox}
\usepackage{caption,subcaption,graphicx}
\usepackage{booktabs}
\usepackage{tikz}
\def\checkmark{\tikz\fill[scale=0.4](0,.35) -- (.25,0) -- (1,.7) -- (.25,.15) -- cycle;}
\usepackage{multirow}
\usepackage{graphicx}

\usepackage{adjustbox}
\usepackage{caption,subcaption,graphicx}

\usepackage{tikz}
\def\checkmark{\tikz\fill[scale=0.4](0,.35) -- (.25,0) -- (1,.7) -- (.25,.15) -- cycle;}
\usepackage{multirow}
\usepackage{graphicx}

\usepackage{enumitem}

\usepackage{ccicons}          
\usepackage{comment}
\usepackage{etoolbox}
\usepackage{lineno}

\usepackage{amsmath}
\DeclareMathOperator*{\argmax}{arg\,max}

\colorlet{RED}{red}

\usepackage[normalem]{ulem}

\begin{document}

\title{Design, Development, and Evaluation of an Interactive Personalized Social Robot to Monitor and Coach Post-Stroke Rehabilitation Exercises 
}

\titlerunning{Interactive Personalized Social Robot to Monitor \& Coach Rehabilitation Exercises}        

\author{Min Hun Lee \and Daniel P. Siewiorek \and \\Asim Smailagic \and
        Alexandre Bernardino\and
        Sergi Bermúdez i Badia
}


\institute{Min Hun Lee \at
           Singapore Management University \\
              \email{mhlee@smu.edu.sg}           
           \and
           Daniel P. Siewiorek, Asim Smailagic \at
           Carnegie Mellon University \\
           \email{\{dps, asim\}@cs.cmu.edu}           
           \and
           Alexandre Bernardino \at
              Instituto Superior Técnico \\
          \email{alex@isr.tecnico.ulisboa.pt}
          \and
          Sergi Bermúdez i Badia \at
             University of Madeira, NOVA-LINCS\\
              \email{sergi.bermudez@uma.pt}
}

\date{Published: March 11, 2023}

\maketitle

\begin{abstract}
Socially assistive robots are increasingly being explored to improve the engagement of older adults and people with disability in health and well-being-related exercises. However, even if people have various physical conditions, most prior work on social robot exercise coaching systems has utilized generic, predefined feedback. The deployment of these systems still remains a challenge. In this paper, we present our work of iteratively engaging therapists and post-stroke survivors to design, develop, and evaluate a social robot exercise coaching system for personalized rehabilitation. Through interviews with therapists, we designed how this system interacts with the user and then developed an interactive social robot exercise coaching system. This system integrates a neural network model with a rule-based model to automatically monitor and assess patients' rehabilitation exercises and can be tuned with individual patient's data to generate real-time, personalized corrective feedback for improvement. With the dataset of rehabilitation exercises from 15 post-stroke survivors, we demonstrated our system significantly improves its performance to assess patients' exercises while tuning with held-out patient's data. In addition, our real-world evaluation study showed that our system can adapt to new participants and achieved 0.81 average performance to assess their exercises, which is comparable to the experts' agreement level. We further discuss the potential benefits and limitations of our system in practice.
\end{abstract}

\input{contents/structure}

\begin{acknowledgements}
The authors thank all the participants in this study for their time and valuable inputs. This work was partially supported by the National Science Foundation (NSF) under grant number CNS-1518865. Additional support was provided by the IntelligentCare project (LISBOA-01-0247-FEDER-045948), the FCT LARSyS funding 2020-2023 (UIDB/50009/2020), the FCT project HAVATAR (PTDC/EEI-ROB/1155/2020), and the Singapore Ministry of Education (MOE) Academic Research Fund (AcRF) Tier 1 grant.
\end{acknowledgements}

%
\section*{Conflict of interest}
The authors have no conflicts of interest to declare that are relevant to the content of this article.

\bibliographystyle{spmpsci}      
\bibliography{main}   

\input{contents/appendix}

\end{document}

%% file: contents/structure.tex
\input{contents/intro.tex}
\input{contents/related.tex}

\input{contents/designs-stroke}
\input{contents/methods.tex}
\input{contents/experiments-system.tex}
\input{contents/results-system.tex}
\input{contents/discussion}
\input{contents/conclusion}

%% file: contents/intro.tex
\section{Introduction}
As the world's older population continues to grow at an unprecedented rate, the current supply of care providers is insufficient to meet the current and ongoing demand for care services \cite{dall2013aging}. Researchers have explored an opportunity of socially assistive robots \cite{feil2005defining,tapus2006towards} that aim to enable people with cognitive, sensory, and motor impairments or assist the clinical workforce \cite{riek2017healthcare}. One potential application is socially assistive robots for rehabilitation therapy \cite{mataric2007socially,lee2020towards,lee2022enabling}. During rehabilitation, patients require completing a significant amount of self-directed exercises \cite{o2019physical,lee2022enabling}. However, low treatment adherence is a problem across several healthcare disciplines of physiotherapy \cite{kaaringen2011elderly}.
To address these problems, there has been increasing attention on  social robot coaching systems \cite{riek2017healthcare,mataric2007socially,lee2020towards,lee2022enabling}. These systems autonomously monitor patients' exercises and provide encouragement to support patients' engagement in well-being related or rehabilitation exercises through social interaction \cite{tapus2007grand,feil2005defining}. 

Prior work on robotic exercise coaching systems has demonstrated that older adults or post-stroke subjects can successfully exercise and stay engaged with a robot over sessions \cite{fasola2013socially,gorer2013robotic}. However, despite of the potential of a robot to monitor and guide exercises, prior work is limited to providing generic, pre-defined corrective feedback on patient's exercise performance (e.g. checking angular difference with the pre-specified motion \cite{gorer2013robotic,fasola2013socially,guneysu2017socially}). It is still challenging to empower a social robot exercise coaching system to generate tailored corrective feedback on an individual patient's motion \cite{gorer2013robotic}, and adopt these systems broadly yet.

\begin{figure*}[h!]
\centering
\begin{subfigure}[t]{.64\textwidth}
\centering
  \includegraphics[width=1.0\columnwidth]{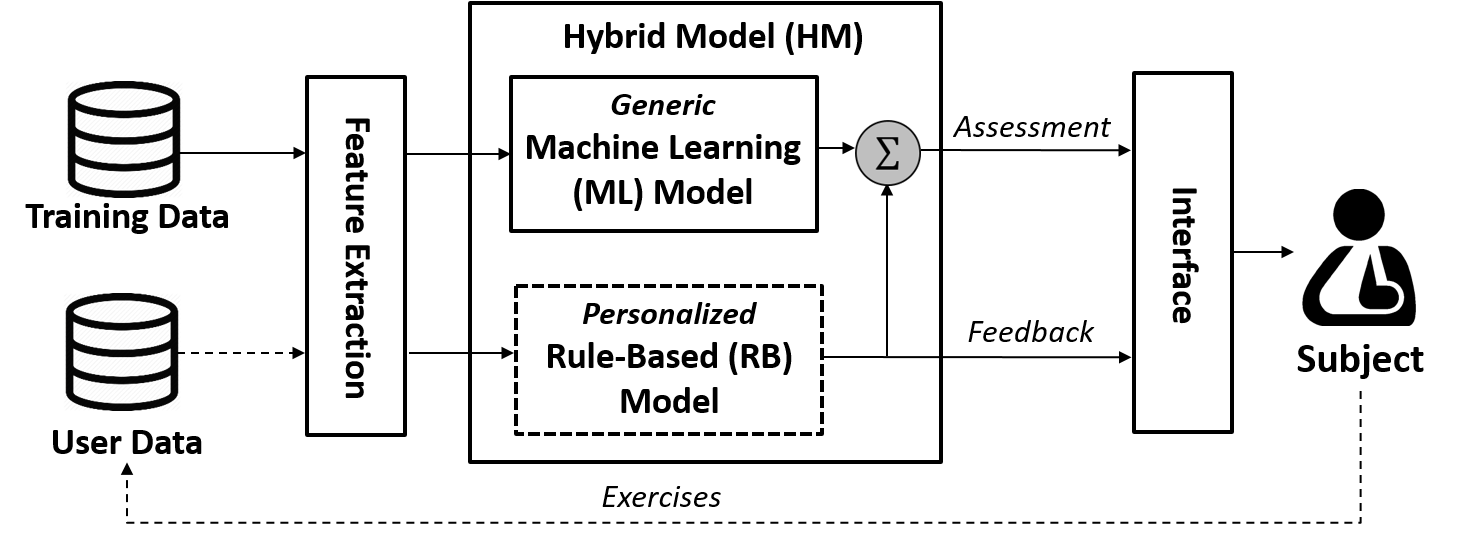}
  \caption{}
  \label{fig:flow-diagram}
\end{subfigure}
\begin{subfigure}[t]{.35\textwidth}
  \centering
   \includegraphics[width=1.0\columnwidth]{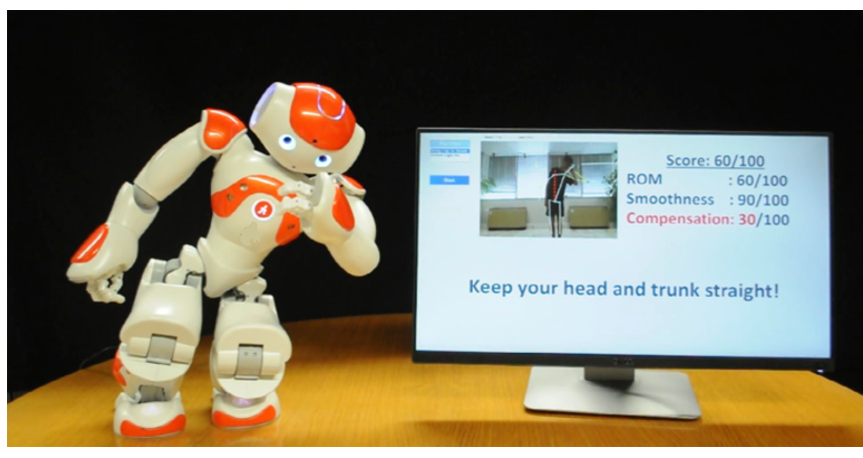}
  \caption{}
  \label{fig:setup}
\end{subfigure}
    \caption{(a) Flow diagram of an interactive approach of a socially assistive robot for personalized physical therapy. (b) a setup of the system with a visualization interface and a socially assistive robot that provides corrective feedback (e.g. audio, visual, gestures of the robot)}
\end{figure*}

In this work, we design, develop, and evaluate a socially assistive robot coaching system that automatically monitors and coaches physical rehabilitation therapy. Specifically, we selected a test domain as stroke, which is the second leading cause of death and disability \cite{feigin2017global}. We then conducted interviews with therapists to design and develop a socially assistive robot coaching system. This system integrates a machine learning (ML) model with a rule-based (RB) model and can be tuned with held-out user data to assess the performance of exercises for personalized post-stroke therapy (Figure \ref{fig:flow-diagram}) \cite{lee2020towards}. Building upon the previous work \cite{lee2020towards}, we demonstrated the benefit of our approach to adapt a new user and provide personalized assessment compared to the commonly used transfer learning technique on a feed-forward neural network model \cite{zhuang2020comprehensive} (i.e. pre-trains a model using the dataset from post-stroke survivors and then finetune it based on the data from a new post-stroke survivor).

During the real-world study with ten participants, our interactive system can be adapted to new participants and achieved 0.81 average performance to assess participants' quality of motion, which is comparable to experts' agreement level (i.e. 0.80 average performance). Overall, participants expressed positive opinions on our system to monitor and provide feedback on their exercises, but also described practical issues to be improved.

 \label{fig:approach}

%% file: contents/related.tex
\section{Related Work}\label{sect:related}
In this section, we describe the background of socially assistive robotics for coaching exercises and outline related work on designs and techniques of socially assistive robotics for rehabilitation therapy. 

\subsection{A Socially Assistive Robot as a Coach}
The research of socially assistive robotics has shown great potential to supplement healthcare services through social interaction \cite{feil2005defining,tapus2009use,mataric2016socially}. For instance, researchers have explored the feasibility of a socially assistive robot exercise coaching system in a rehabilitation process, in which the system automatically monitors rehabilitation exercises and provides users feedback without the presence of a therapist \cite{mataric2007socially}. Fasola and Mataric demonstrated that older adults considered a physically embodied robot more engaging and acceptable as an exercise partner than a virtually embodied agent \cite{fasola2013socially}. Furthermore, researchers have shown that diverse populations (i.e. post-stroke patients \cite{fasola2013socially}, elderly people \cite{gorer2017autonomous}, children \cite{guneysu2017socially}) can engage in exercise sessions with a social robot exercise coaching system on several domains (e.g. stroke, dementia, etc.) \cite{tapus2009use,lee2022enabling,riek2017healthcare}. 

\subsection{Designs of a Socially Assistive Robot Coaching System}
For creating user-centered, socially assistive robots, researchers have engaged with various stakeholders to derive design requirements \cite{beer2012domesticated,azenkot2016enabling,winkle2018social,lee2022enabling}.
Winkle et al. described design guidelines of social robots for rehabilitation, from focus group sessions and interviews with therapists \cite{winkle2018social}. In addition, Polak and Levy-Tzedek also conducted focus group sessions with therapists and a preliminary evaluation study on a gamification system for rehabilitation with four post-stroke survivors \cite{feingold2020social}. Lee et al. conducted studies with therapists and post-stroke survivors to elicit detailed design specifications on how AI and robotic coaches could interact with and guide patients' exercises in an effective and acceptable way \cite{lee2022enabling}. One of the important design considerations that are repetitively mentioned in prior work is the importance of personalized feedback \cite{winkle2018social,feingold2020social,lee2022enabling}. 

In this work, we interviewed therapists to understand what kinds of feedback they generate and explored a computational technique that enables a social robot exercise coaching system to generate personalized feedback and control robot behaviors as a therapist. 

\subsection{Techniques of Monitoring and Assessing Patient's Exercises}
The capability of automatically assessing a patient's motion and providing a personalized interaction with tailored corrective feedback on patient's exercise performance is critical for the deployment of a social robot exercise coaching system \cite{mataric2009socially,gorer2017autonomous,lee2022enabling}. For personalized interactions with a socially assistive robot, Irfan et al. explored recognizing a user and refer the user's name periodically as personalized feedback \cite{irfan2020using}. Schneider and Kummert investigated a technique to match the user's preferred order of different exercises for personalized interactions of exercise robots \cite{schneider2021comparing}. However, limited prior work on social robot exercise coaching systems has explored how an automated assessment approach can be developed to generate personalized corrective feedback.

When it comes to an automated assessment approach, researchers have implemented a method that monitors the completion of an exercise by computing the difference of a joint angle between the user's motion and the pre-defined target motion \cite{fasola2013socially,gorer2017autonomous}. Guneysu and Arnrich \cite{guneysu2017socially} applied dynamic time warping to compute the statistics of a joint angle and distance measures with a pre-defined motion. Nguyen et al. \cite{tanguy2016computational} utilized a Gaussian Mixture Model to generate an ideal motion and arbitrarily set a threshold value to identify the differences of joints between idea and observed motions. 
Although both \cite{guneysu2017socially} and \cite{tanguy2016computational} support analyzing multiple variables for evaluating an exercise, they still rely on a pre-defined motion or a generic threshold. Prior work with generic threshold-based methods might not be applicable for patients with various characteristics \cite{lee2020study}. 

In addition, researchers have also explored a machine learning model to monitor patients' quality of motion. For instance, Kashi et al. evaluated the feasibility of a random-forest model to identify compensatory movements \cite{kashi2020machine}. However, it remains unclear how such a machine learning model can adapt to a new patient and perform well to assess patients' quality of motion. 

For personalized quantitative rehabilitation assessment, Lee et al. explored an approach of dynamic feature selections \cite{lee2020designing} and a hybrid model that integrates a machine learning model with a rule-based model \cite{lee2020interactive}. However, prior work is limited to providing assessment after completing a motion and does not support frame-level assessment to provide any information on when an erroneous motion has occurred.

Building upon prior work that explores a hybrid model for personalized assessment \cite{lee2020interactive,lee2020towards}, we further investigated the system implementation of a socially assistive robot to automatically monitor and guide patients' exercises. Specifically, we analyzed the benefit of our interactive hybrid approach compared to the commonly used transfer learning technique on a feed-forward neural network model \cite{zhuang2020comprehensive,weiss2016survey} (i.e. pre-trains the model using the dataset from other post-stroke survivors and then finetune it based on data from a new post-stroke survivor). In addition, we conducted a real-world experiment to evaluate the feasibility to adapt to a new participant and provide personalized, real-time corrective feedback.

%% file: contents/designs-stroke.tex
\section{Study for Stroke Rehabilitation} \label{sect:study-design}
This work focuses on the domain of stroke, which is the second leading cause of death and third most common contributor to disability \cite{feigin2017global}.
First, we iteratively discussed with three therapists (TPs with check marks in the specification column of Table \ref{tab:hri-study-design-tps}; mean (M) $= 6.33$, standard deviation (SD) $= 2.05$ years of experience in stroke rehabilitation) to specify the study designs on stroke rehabilitation: exercises and performance components for assessment \cite{lee2019learning}. 
We then had additional interviews with therapists to learn their practices on how they guide rehabilitation assessment. Based on these interviews, we created an interactive social robot coaching system that automatically monitors and coaches rehabilitation exercises. We then conducted a real-world experiment with ten healthy participants to evaluate the potential benefits and limitations of our system.
This section presents only the specifications of our study and interviews with therapists to understand their practices. The evaluation part will be discussed later in Section \ref{sect:results-therapists} after presenting our system implementation.

\begin{table}[htp]
\centering
\caption{Profiles of therapists, who contributed to specify the study and share their practices to design our system}
\label{tab:hri-study-design-tps}
\resizebox{0.6\textwidth}{!}{%
\begin{tabular}{@{}cccc@{}}
\toprule
ID &
  \begin{tabular}[c]{@{}c@{}}Specification\end{tabular} & \begin{tabular}[c]{@{}c@{}}Interview\end{tabular} &
  \begin{tabular}[c]{@{}c@{}}\# of Years in\\ Stroke Rehabilitation\end{tabular} \\ \midrule
TP 1 & \checkmark & \checkmark & 6  \\
TP 2 & \checkmark & \checkmark & 4  \\
TP 3 & \checkmark & & 9  \\
TP 4 &   & \checkmark & 23 \\ \bottomrule
\end{tabular}%
}
\end{table}

\subsection{Three Task-Oriented Upper Limb Exercises}
This work utilizes three upper-limb stroke rehabilitation exercises recommended by therapists \cite{lee2020towards}. For Exercise 1, a user has to raise the user's wrist to the mouth as if drinking water (Figure \ref{fig:sample-compensation-normal-e1-p14}). For Exercise 2, a user has to raise the user's wrist forward as if touching a light switch on the wall (Figure \ref{fig:sample-compensation-normal-e2-p8}). For Exercise 3, a user has to extend the user's elbow in the seated position to practice the usage of a cane (Figure \ref{fig:sample-compensation-normal-e3-p14}).

\subsection{Unaffected and Affected Sides}
When a stroke occurs, post-stroke survivors suffer from the paralyzed, limited functional abilities of limbs. In this work, we refer to the unparalyzed side of a post-stroke survivor as the \textit{``Unaffected''} side and the paralyzed side of a post-stroke survivor with limited functional ability as the \textit{``Affected''} side. 

\subsection{Performance Components}
\begin{figure}[tp!]
\centering
\begin{subfigure}[t]{.25\columnwidth}
\centering
  \includegraphics[width=.68\columnwidth]{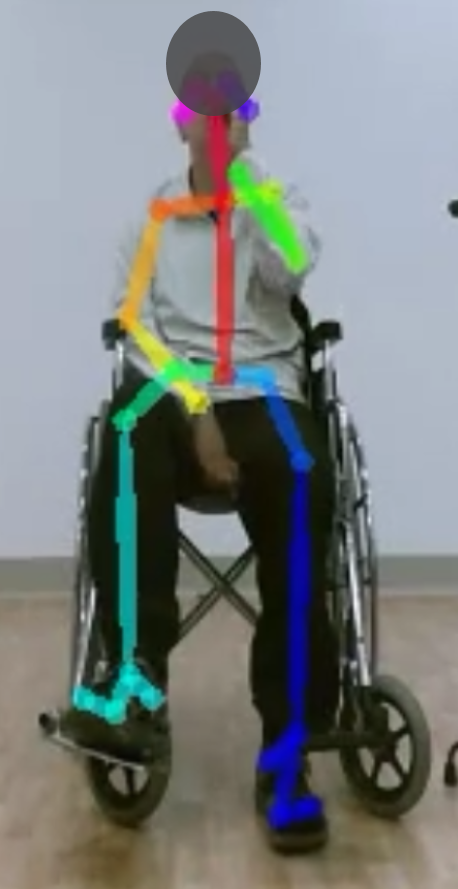}
  \caption{E1-Unaffected}
  \label{fig:sample-compensation-normal-e1-p14}
\end{subfigure}\hfill  
\begin{subfigure}[t]{.25\columnwidth}
\centering
  \includegraphics[width=.69\columnwidth]{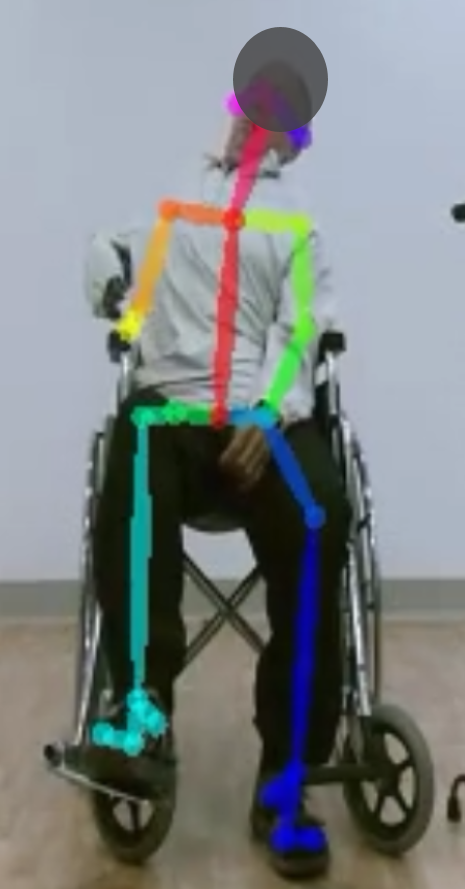}
  \caption{E1-Affected}
  \label{fig:sample-compensation-abnormal-e1-p14}
\end{subfigure}\hfill
\begin{subfigure}[t]{.25\columnwidth}
\centering
  \includegraphics[width=.68\columnwidth]{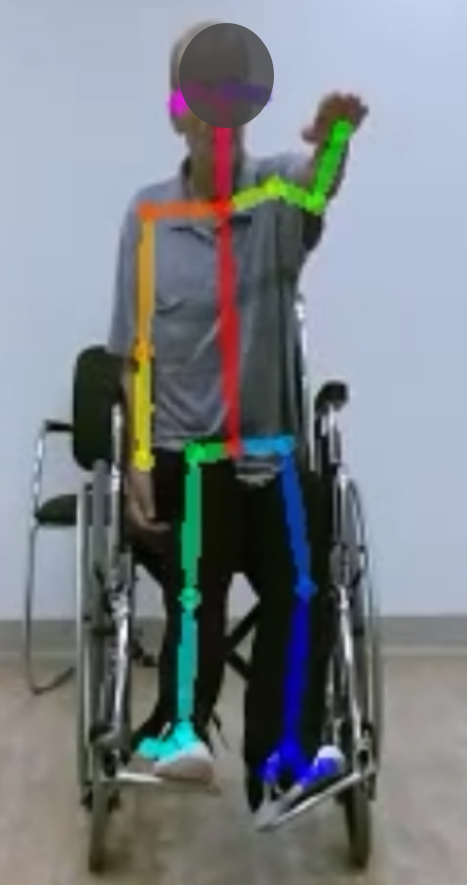}
  \caption{E2-Unaffected}
  \label{fig:sample-compensation-normal-e2-p8}
\end{subfigure}\hfill  
\begin{subfigure}[t]{.25\columnwidth}
\centering
  \includegraphics[width=.69\columnwidth]{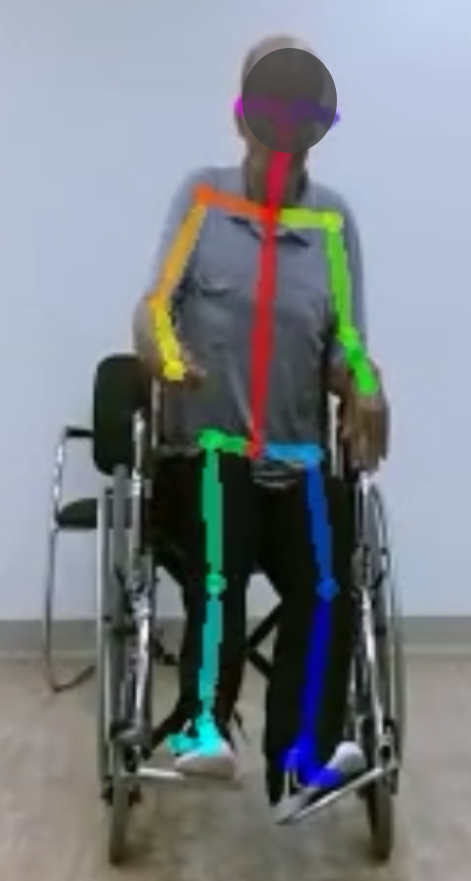}
  \caption{E2-Affected}
  \label{fig:sample-compensation-abnormal-e2-p8}
\end{subfigure}\hfill  
\begin{subfigure}[t]{.25\columnwidth}
\centering
  \includegraphics[width=.68\columnwidth]{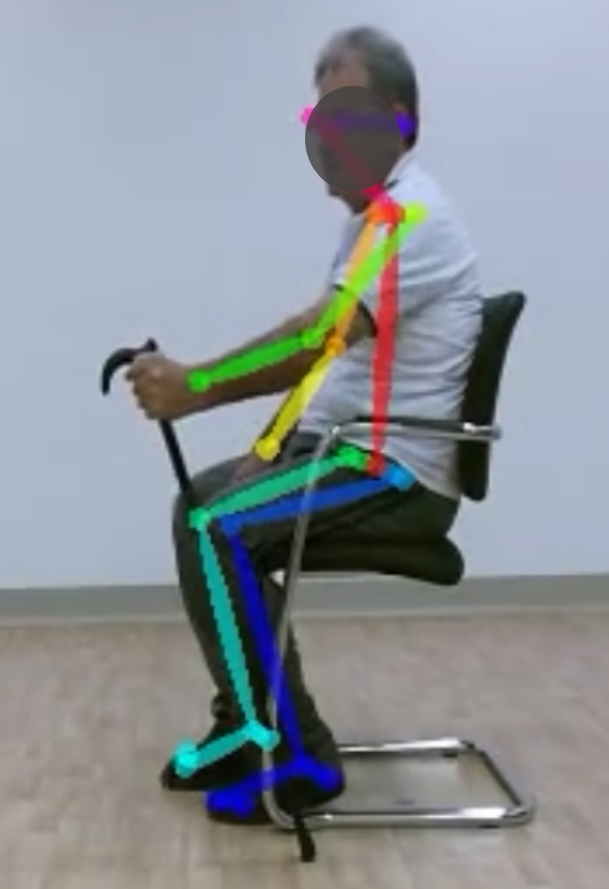}
  \caption{E3-Unaffected}
  \label{fig:sample-compensation-normal-e3-p14}
\end{subfigure}  
\begin{subfigure}[t]{.25\columnwidth}
\centering
  \includegraphics[width=.69\columnwidth]{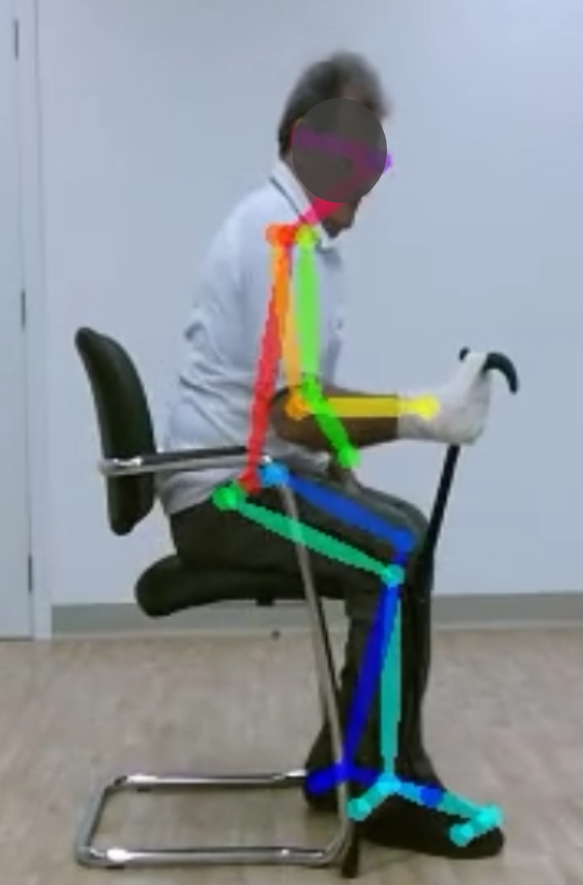}
  \caption{E3-Affected}
  \label{fig:sample-compensation-abnormal-e3-p14}
\end{subfigure}
\caption{Sample unaffected and affected motions of exercises: (a) a patient raises the patient's unaffected side of the wrist to the mouth, (b) a patient compensated with trunk and shoulder joints when attempting to move the patient's affected side of the wrist, (c) a patient raises patient's unaffected side of the wrist forward, (d) a patient elevated shoulder to compensate the limited functional ability of the patient's affected side, (e) a patient extends the patient's affected side of the elbow, and (f) a patient leaned trunk forward to extend the patient's elbow.}~\label{fig:sample-compensation}
\end{figure}

We discussed commonly used stroke assessment tools (i.e. the Wolf Motor Function Test \cite{wolf2001assessing} and the Fugl Meyer Assessment \cite{sanford1993reliability}) with therapists and specified three common performance components of stroke rehabilitation exercises: \textit{`Range of Motion (ROM)'}, \textit{`Smoothness'}, and \textit{`Compensation'} \cite{lee2020towards}. 
The \textit{`ROM'} indicates how closely a patient performs the target position of a task-oriented exercise. The \textit{`Smoothness'} describes the degree of trembling and irregular movement of joints while performing an exercise. The \textit{`Compensation'} indicates whether a patient performs any unnecessary, compensatory movements to achieve a target movement. For instance, patients might lean their head or trunk to the side and elevate their shoulder to perform an exercise using their affected side with the limited functional ability (Figure \ref{fig:sample-compensation}).

The descriptions and labels of performance components are described in Table  \ref{tab:score-guidelines}. The labels of \textit{`ROM'} and \textit{`Smoothness'} are annotated at the end of a motion and represented as a binary label on each performance component: a correct/normal performance component ($Y = 1$) and an incorrect/abnormal performance component ($Y = 0$). The labels of \textit{`Compensation'} are annotated at every frame of the patient's motion to indicate whether three major compensations (i.e. abnormal alignment of head, spine, and shoulder) occur or not.

\input{tables/score-guidelines}

\subsection{Understanding Therapists' Practices}\label{sect:study-practices}
We conducted a one-on-one interview with each of the three therapists (TPs with check marks in the interview column of Table \ref{tab:hri-study-design-tps}; mean (M) $= 11.00$, standard deviation (SD) $= 8.52$ years of experience in stroke rehabilitation). During the one-hour interview, the researcher asked therapists to speak aloud their strategies for coaching a rehabilitation session and providing feedback on patient's exercises (i.e. \textit{``what kinds of feedback do you generate for a post-stroke survivor?''}). To assist therapists' speaking aloud process, the researcher showed them the videos of post-stroke survivors, who have different functional abilities (i.e. high, moderate, and low capability to achieve an exercise) and perform rehabilitation exercises. The detailed process of collecting these videos is described in Section \ref{sect:dataset}. 

We analyzed the transcripts of interviews with therapists through an iterative coding process \cite{gale2013using}. Specifically, we first open coded interview transcripts, discussed emerging themes and ideas, and iteratively improved our codebook. We found that therapists oversee the treatments of a post-stroke survivor by providing a personalized rehabilitation session. Specifically, they monitor how their patients perform an exercise and provide their patients feedback to support the correct execution of an exercise and encourage participation in rehabilitation \cite{lee2022enabling}. For guiding a rehabilitation session, we noticed that all three therapists have a simple and common procedure \cite{lee2022enabling}. Specifically, when they start a session, they engage with their patients through brief greetings and describe the goal of a session (e.g. what kinds of exercises a patient will perform and how many repetitions are recommended) \cite{lee2022enabling}. If a patient is not familiar with an exercise motion, therapists might show themselves to instruct a motion that a patient has to practice \cite{lee2022enabling}. When a patient performs an exercise, therapists monitor the patient's exercises to identify any part for improvement and provide corrective feedback \cite{lee2022enabling}. For instance, we found that therapists are particularly attentive to providing feedback on compensatory motions \cite{cirstea2000compensatory} that might cause more severe pains. As rehabilitation requires patient engagement over an extended period, therapists also strive to provide positive encouragement to their patients \cite{lee2022enabling}. 

%% file: tables/score-guidelines.tex
\begin{table}[thp!]
\centering
\caption{Performance Components and Labels of Physical Stroke Rehabilitation Exercises.}
\resizebox{1.0\columnwidth}{!}{%
\begin{tabular}{ccl} \toprule
\textbf{\begin{tabular}[c]{@{}c@{}}Performance \\ Components\end{tabular}} & \textbf{\begin{tabular}[c]{@{}c@{}}Labels\end{tabular}} & \multicolumn{1}{c}{\textbf{Guidelines}} \\  \midrule
\multirow{2}{*}{\begin{tabular}[c]{@{}c@{}}Range of Motion \\ (ROM) 
\end{tabular}
} & 0 & Movement that does not achieve a \textit{`Target'} position
 \\
 & 1 & Movement achieves a \textit{`Target'} position \\ \midrule
\multirow{2}{*}{Smoothness} & 0 & Movement with tremor or unsmooth coordination \\
 & 1 & Smoothly coordinated movement \\ \midrule
\multirow{3}{*}{Compensation} & 0/1 & Head in abnormal/normal alignment \\
 & 0/1 & Spine in abnormal/normal alignment\\
 & 0/1 & Shoulder in abnormal/normal alignment\\ \bottomrule
\end{tabular}%
}
\label{tab:score-guidelines}
\end{table}

%% file: contents/methods.tex
\section{Interactive Approach of an Socially Assistive Robot for Personalized Assessment and Feedback}\label{sect:method}
This work presents an interactive approach of a social robot exercise coaching system (Figure \ref{fig:flow-diagram}), which combines machine learning (ML) and rule-based (RB) models to assess the performance of patient's exercises and tunes with patient's data to generate personalized feedback \cite{lee2020towards}. An ML model of our approach aims to extract meaningful patterns from a large amount of data and to support a generic assessment of the patient's quality of motion \cite{lee2020study}. As such a generic ML model might not perform well on an unobserved new patient's motion with unique characteristics, our approach also integrates an ML model with a personalized RB model that can tune with the patient's unaffected motions to derive patient-specific threshold values. This RB model can be easily updated to complement a generic ML model through a weighted average, ensemble technique \cite{lee2020study} into a hybrid model (HM) and utilized to generate personalized corrective feedback on patient's exercises. To provide feedback when an erroneous motion has occurred, we explored an ensemble voting method that leverages predictions on multiple consecutive frames for a more accurate frame-level assessment \cite{lee2020towards}. In the following subsections, we describe the components of our approach: feature extraction, ML models, RB models, hybrid models, an ensemble voting method, and an interface of a socially assistive robot for personalized rehabilitation therapy.

\subsection{Feature Extraction}\label{sect:method-feature}
This work represented an exercise motion with sequential joint coordinates from a Kinect v2 sensor (Microsoft, Redmond, USA) and extracted various kinematic features \cite{lee2019learning}. For the \textit{`ROM'} performance component, we computed joint angles (e.g. elbow flexion, shoulder flexion, elbow extension), the distance to a target position, and normalized relative joint trajectories (i.e. the Euclidean distance between two joints - head and wrist, head and elbow) \cite{lee2019learning}. For the \textit{`Smoothness'} performance component, we computed the following speed-related features: speed and the zero-crossing ratio of acceleration  \cite{lee2019learning}. As our work focuses on upper-limb exercises, we computed these speed-related features on wrist and elbow joints. For the \textit{`Compensation'} performance component, we computed normalized joint trajectories: distances between joint positions of the head, spine, and shoulder in $x$, $y$, $z$ axis from the initial to the current frame \cite{lee2019learning}.

A moving average filter with a window size of five frames was applied to reduce the noise of the estimated joint positions from a Kinect sensor similar to \cite{lee2019learning}. Given an exercise motion, we computed a feature matrix $\textbf{F} = \{f_1, ..., f_T\} \in R^{T \times d}$ with $T$ number of frames and $d$ features and the statistics (e.g. maximum, minimum, range, average, and standard deviation) of a feature matrix over all frames of the exercise to summarize a motion into a feature vector, $X \in R^{5d}$. This summarized feature vector was utilized for the assessment of \textit{`ROM'} and \textit{`Smoothness'} performance components. In addition, unlike \cite{lee2019learning} that only supports offline assessment on the \textit{`Compensation'} performance component, we extracted a feature vector at each frame for real-time, frame-level assessment on the \textit{`Compensation'} performance component. Overall, we extracted 30 features for the \textit{`ROM'} performance component, 60 features for the \textit{`Smoothness'} performance component, and 9 features for the \textit{`Compensation'} performance component.

\subsection{Machine Learning (ML) Model}\label{sect:method-ml}
For a  machine learning (ML) model, we applied a supervised learning algorithm through leave-one-patient-out cross-validation that utilizes training data from all patients except a patient for testing. The ML model computes the score of being correct on a performance component ($P_{ML}$) and predicts the quality of motion. Among various supervised learning algorithms, a Decision Tree, Linear Regression, a Support Vector Machine, a feed-forward Neural Network, and a Long Short Term Memory (LSTM) network, we utilized a feed-forward Neural Network (NN) model due to its outperformance as shown in \cite{lee2020towards}. For the implementation of a feed-forward neural network (NN) model, we explored various architectures (i.e. one to three layers with 32, 64, 128, 256, and 512 hidden units) and an adaptive learning rate with different initial learning rates (i.e.
0.0001, 0.005, 0.001, 0.01, 0.1). We applied ‘ReLu’ activation functions and ‘AdamOptimizer’ and trained a model with cross-entropy loss and a mini-batch size of 1 and an epoch of 1.

\subsection{Rule-Based (RB) Model}\label{sect:method-rb}
A rule-based (RB) model leverages the set of feature-based, \textit{if-then} rules from therapists to estimate the quality of a motion \cite{lee2020study}. In addition, our system computes statistics of kinematic features from user data and generates patient-specific rules for personalized assessment.
For the initial development of the RB model, semi-structured interviews were conducted with two therapists (mean (M) $= 5.05$, standard deviation (SD) $= 1.05$ years of experience in stroke rehabilitation) to elicit their knowledge of assessing stroke rehabilitation exercises. The knowledge of therapists has been formalized as 15 independent \textit{if-then} rules (Appendix. Table \ref{tab:method-rules-all}). 
For example, the assessment on the ROM component for Exercise 1 is specified as follows \cite{lee2020study}: 

\begin{equation}
\footnotesize{\hat{Y}} = 
\begin{cases}
  $1$ & \text{if $p^{max}(wr, c_y)$ $>=$ $p^{max}(spsh, c_y)$} \\
  $0$ & \text{else} \\
\end{cases}
\end{equation}

\noindent
where $p(j, c)$ indicates the joint position ($p$) with a joint $j$ (e.g. wrist $(wr)$ and spine shoulder, the top of spine, $(spsh)$) and the coordinate of a joint ($c$) in the set $C \in \{c_x, c_y, c_z\}$. $\hat{Y}$ denotes the predicted label on a performance component.

This rule simply checks the maximum position of a wrist joint, $p^{max}(wr, c_y)$, related to that of a spine shoulder joint, $p^{max}(spsh, c_y)$, in the y-coordinate to roughly estimate whether a patient achieves the target position of Exercise 1. For the prediction with multiple rules, we apply a majority voting algorithm and do not apply any tie-breaking method given an odd number of rules. 

The score of being correct on each performance component using the RB model ($P_{RB}$) is computed with the following equation:

\begin{equation}
P_{RB} = \frac{1}{|\mathbb{R}|}\sum_{r \in R} \min(\frac{{x}_{r}}{{\tau}_{r}}, 1)
\end{equation}

\noindent where 
$x_r$ indicates the feature value of a rule $r$ from a trial (e.g. $p^{max}(wr, c_y)$ for the example above), $\tau_{r}$ describes the threshold value of a rule $r$ (e.g. $p^{max}(spsh, c_y)$ for the example above). $\mathbb{R}$ describes the set of rules elicited from the therapists. $\min$ function is applied so that this equation assigns a value of 1 if the feature value of a rule exceeds the threshold of that rule. Otherwise, the equation normalizes the feature value of a rule with the threshold of a rule to compute the score of being correct.

In addition, as the initial threshold values of rules are generic, our approach can further tune a rule-based (RB) model with held-out user's unaffected motions to update its threshold values on each patient (Figure \ref{fig:flow-diagram}). For the computation of personalized threshold values, we utilize the held-out user's unaffected motions to learn a Gaussian probability density function $f(x_r) \sim N(\mu_r, \sigma_r^2)$. Specifically, when a patient first interacts with the system and there is no pre-stored patient's unaffected data, the system will inform the patient to perform an exercise with the patient's unaffected side. When the system has the patient's unaffected data, it will process to extract the feature value of a rule ($x_r$). We then utilized the maximum likelihood estimate (MLE) \cite{gopinath1998maximum} to estimate the parameters of a Gaussian probability density function, $\mu_r$ and $\sigma_r$  as the mean and standard deviation of $x_r$ respectively. We then update the threshold value for a rule $r$ with either $2\sigma_s$ or $3\sigma_s$  (i.e. $\tau_r \in [\mu_r + 2\sigma_r, \mu_r + 3\sigma_r]$).

\subsection{Hybrid Model}
A hybrid model (HM) applies a weighted average, ensemble technique \cite{baltruvsaitis2019multimodal} to combine machine learning (ML) and rule-based models to assess patients' quality of motion \cite{lee2020study}. 
For the prediction on the quality of motion, the HM computes the weighted average of prediction scores from two models, in which the contribution weight of each model is the performance of a model (i.e. the F1-score of each model in the range of $[0, 1]$). Given training data, this weight can be pre-computed and updated as the system collects additional data.
The equation of computing the score of being correct using the HM, $P_{HM}$ is as follows:

\begin{equation}
P_{HM} = \frac{\rho_{ML}}{\rho_{ML} + \rho_{RB}}P_{ML} + \frac{\rho_{RB}}{\rho_{ML} + \rho_{RB}}P_{RB}
\end{equation}

\noindent
where $P_{ML}$ and $P_{RB}$ indicate the scores of the machine learning (ML) and rule-based (RB) models, and $\rho_{ML}$ and $\rho_{RB}$ describe the weights, F1-scores of ML and RB models.

\subsection{Ensemble Voting Method for Frame-Level Assessment}
Our approach can detect a compensation motion at the frame level in real-time so that a social robot exercise coaching system can provide a patient with corrective feedback when an erroneous motion has occurred. However, such a frame-level assessment, identifying the exact boundaries of a compensation motion is challenging \cite{hasan2014continuous}. Thus, our approach applies an ensemble voting method \cite{dietterich2000ensemble} that utilizes predictions on multiple consecutive $V_f$ frames for a more robust frame-level assessment. The process of this method  consists of 1) initial, continuous frame-level predictions and 2) the computation of a score to determine a winning prediction. 

Let us denote $h(f_t)$ the predicted frame-level assessment at frame $t$ with an assessment model $h$ (e.g. a machine learning model, a rule-based model, or a hybrid model) and a feature vector $f_t$. 
The first process of an initial frame-level prediction runs continuously with an assessment model to generate predicted frame-level assessment $h(f_t)$ at each frame $t$.
When $V_f$ number of initial frame-level predictions are available, our method computes a score of detecting a compensation motion at frame $t$ over all label classes $Y \in \mathcal{Y}$. Then, the winning prediction at frame $t$ is selected as follows:

\begin{equation}
    \hat{Y}_t = \argmax_{Y\in \mathcal{Y}} \sum_{f_t \in \bar{F}} \delta(h(f_t), Y)
\end{equation}

\noindent where $\bar{F}$ indicates a set of accumulated $V_f$ feature vectors until $t$ frame and $\delta(h(f_t), Y)$ assigns 1 if $h(f_t) = Y$ and 0 otherwise. The $\delta$ function is to count the predicted assessment of $Y$ with the predictions from $V_f$ frames. $\hat{Y}_t$ indicates the predicted frame-level assessment at $t$ frame on a compensation motion with the largest number of the predictions, votes from $V_f$ frames. In case of having tied votes, our method assigns $\hat{Y}_t$ with the latest prediction $h(f_t)$. 
By leveraging votes from past $V_f - 1$ frames to the current $t$ frame, our approach can support a more robust frame-level assessment. 

\subsection{Interface of a Socially Assistive Robot}
\begin{figure}[h!]
\centering
\resizebox{0.8\columnwidth}{!}{%
  \includegraphics[width=1.0\columnwidth]{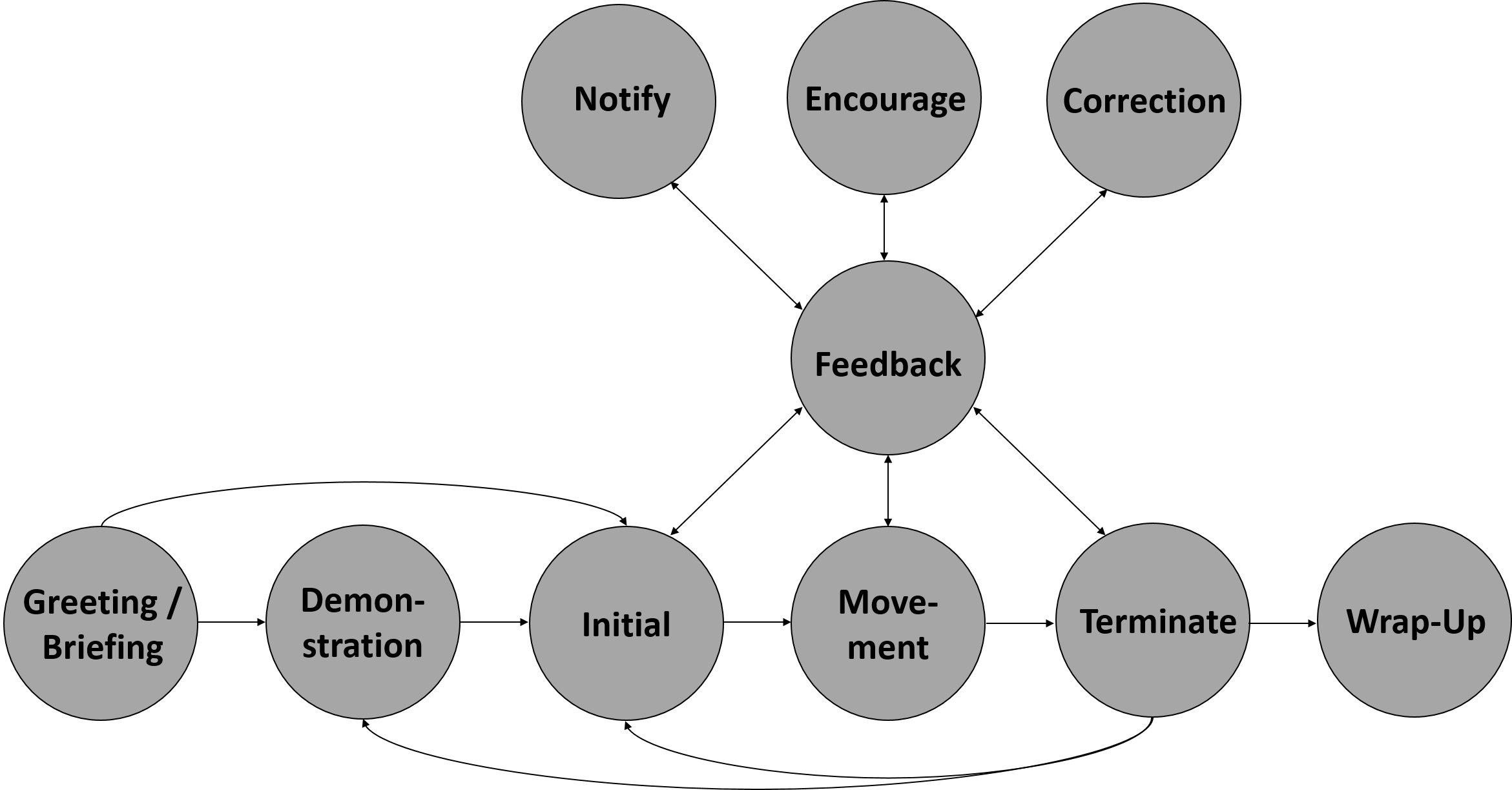}
  }
  \caption{The state machine of an interactive, social robot exercise coaching system: The system will provide various types of social interactions, such as audio and visual feedback to a user and control the gestures of the social robot based on each state (e.g. greeting, demonstration, feedback on exercise, and wrapping up a session)}
  \label{fig:state-diagram}
\end{figure}

Based on our findings from the interviews with therapists (Section \ref{sect:study-practices}), we designed and developed a state machine to enable  interactions of our social robot exercise coach system with users as a therapist. This state machine (Figure \ref{fig:state-diagram}) includes ten states: \textit{`Greeting/Briefing'}, \textit{`Demonstration'}, \textit{`Initial'}, \textit{`Movement'}, \textit{`Terminate'}, \textit{`Feedback'}, \textit{`Notify'}, \textit{`Encourage'}, \textit{`Correction'}, and \textit{`Wrap-up'}. Depending on the user inputs (e.g. clicking a button to start a system) and the results from our motion analysis component, the state machine will transit to a corresponding state and generate audio and visual feedback and control the behaviors of our social robot exercise coaching system (e.g. gestures). 

In the \textit{`Greeting/Briefing'} state, our robotic exercise coaching system will summarize the main goal of a rehabilitation session as specified by a therapist. The system will show the video of a prescribed motion in the \textit{`Demonstration'} state if a new exercise is prescribed and a user requests it. In the \textit{`Initial'} state, the system will prompt whether a user is ready to start an exercise. Once a user confirms to start performing an exercise, the system will transit to the \textit{`Movement'} state and alert that the system starts monitoring in the \textit{`Notify'} state. When a user performs an exercise, the system will provide various types of feedback in the \textit{`Feedback'} state. For instance, if the system detects any compensated motion in real-time, the system will provide a user corrective feedback on which unnecessary joints are involved in the \textit{`Correction'} state. Once a user completes an exercise, the state machine of our system transits to the \textit{`Terminate'} state, in which it will summarize the predicted assessment on the quality of motion in the \textit{`Notify'} state and provides \textit{`Encouragement'}. When a user completes all prescribed exercises or requests to finish a session, the system will summarize what a user achieves in the session and remind the next session in the \textit{`Wrap-Up'} state.

For a social assistive robot, we used an NAO robot (SoftBank Robotics Europe, France) that supports competitive hardware capabilities and a user-friendly software development environment with cost reduction \cite{gouaillier2008nao}. We utilized the NAO SDK and Choregraphe software \cite{pot2009choregraphe} to program the gestures of the NAO and Google Text To Speech (TTS) APIs \cite{kepuska2017comparing} to generate audio outputs from a socially assistive robot exercise coaching system. For communicating the results of motion analysis, we implemented client/server applications with socket programming in Python. 

%% file: contents/experiments-system.tex
\section{Experiments}

\subsection{Dataset of Three Upper-Limb Exercises}\label{sect:dataset}
To evaluate the feasibility of our approach, this work utilizes the dataset of three exercises from 15 post-stroke subjects using a Kinect v2 sensor (Microsoft, Redmond, USA) \cite{lee2019learning}. Fifteen post-stroke patients (2 females) with diverse functional abilities from mild to severe impairment (37 $\pm$ 21 out of 66 Fugl Meyer Scores \cite{sanford1993reliability}) performed 10 repetitions of each exercise with both affected and unaffected sides. During the data collection, a sensor was located at a height of 0.72m above the floor and 2.5m away from a subject and recorded the trajectory of joints and video frames at 30 Hz. The starting and ending frames of exercise movements were manually annotated. 

Two therapists (mean (M) $= 5.0$, standard deviation (SD) $= 1.0$ years of experience in stroke rehabilitation) annotated the dataset to implement our approach and compute the experts' agreement level. They individually watched the recorded videos of patients' exercise movements and annotated the performance components of the exercise motion dataset. For the frame-level annotation of the \textit{`Compensation'} performance component, two expert annotators reviewed the images that were extracted from the recorded videos with the corresponding sampling frequency using the FFmpeg tool \cite{developers2016ffmpeg}. The annotations of experts were compared to measure the experts' agreement on F1-scores (i.e. \textit{`Experts' Agreement'} in Appendix. Table \ref{tab:results-avg}, with the Cohen’s Kappa of 0.65). We utilized the annotation of an expert, who evaluated the functional abilities of patients with the Fugl Meyer Assessment and had more experience as the ground truth. 

The collected dataset was divided into \textit{`Training'} and  \textit{`User'} data as follows:

\begin{itemize}[leftmargin=*]
    \item \textbf{\textit{`Training Data'}} (Figure \ref{fig:flow-diagram}) is composed of 140 unaffected motions and 140 affected motions from 14 post-stroke subjects to train a machine learning (ML) model. 
    \item \textbf{\textit{`User Data'}} (Figure \ref{fig:flow-diagram}) includes 10 unaffected motions and 10 affected motions of a testing post-stroke subject. 
\end{itemize}

\subsection{Quantitative In-Lab System Evaluation}
We applied Leave-One-Subject-Out (LOSO) cross-validation on post-stroke patients to evaluate our approach. A machine learning model (ML) was trained with data from all subjects except one testing post-stroke survivor. An initial rule-based (RB) model was developed from the interviews with therapists. A hybrid model applies a weighted average to integrate a trained, outperforming ML model with a rule-based model. All models (e.g. rule-based, machine learning, hybrid) were tested with affected motions of the left-out post-stroke patient. This process was repeated over all post-stroke survivors to evaluate the performance of a model. In addition, we analyzed the effect of tuning a model with held-out unaffected motions of the left-out post-stroke survivor. For a feed-forward neural network model, we applied the common transfer learning technique \cite{zhuang2020comprehensive} that finetunes a pre-trained model with the patient's unaffected motions to implement the tuned feed-forward neural network (Tuned ML-NN). We then compared the performance of the Tuned ML-NN with that of the HM-Tuned to evaluate the value of our interactive HM for a personalized assessment. We also explored different numbers of multiple consecutive $V_f$ frames on our ensemble voting method for frame-level assessment (i.e. $V_f = 1, ..., 30$). For the performance metric, this work utilized an F1-score that computes the harmonic mean of precision and recall for a more realistic measure of a model. 

\subsection{Real-World System Evaluation}
After developing our system, we conducted a real-world experiment to evaluate the potential of our system with healthy participants. 

As we had difficulty with running a study with post-stroke patients due to COVID-19, we aimed to conduct a pilot evaluation study to receive early feedback on our system before conducting user studies with post-stroke survivors. For this real-world evaluation, we recruited 10 healthy participants. In each session, the researcher gave an introduction to the study and instructed a compensatory motion of a post-stroke survivor by showing a video and an image of a post-stroke survivor. When a participant became familiar with a compensatory motion, the researcher instructed the participant to perform six repetitions of an exercise: one trial of a correct \textit{`ROM'} and no \textit{`Compensation'}, one trial of an incorrect \textit{`ROM'} and no \textit{`Compensation'}, two trials of a correct \textit{`ROM'} and acted-out \textit{`Compensation'}, and two trials of an incorrect \textit{`ROM'} and acted-out \textit{`Compensation'}. While performing an exercise, our system automatically monitored the participants' exercises and provided real-time feedback through audio, visualization, and robot gestures (Figure \ref{fig:setup}). All sessions were video-recorded for further analysis (e.g. collecting the ground truth). After completing the exercise trials, each participant filled out the following usability questionnaires  \cite{fasola2013socially} of our system on a 7-point scale and provide any suggestions for improvement:

\begin{itemize}
    \item Usefulness: \textit{``The system provides a useful, valuable, rich feedback''}
    \item Intelligence: \textit{``The system is intelligent and competent''}
    \item Trustfulness: \textit{``The system is trustful''}
    \item Social Attraction: \textit{``The system is friendly and pleasant. I could have an enjoyable and motivating interaction''}
    \item Usage Intention: \textit{``I would use the system in future or recommend the system as an exercise partner''}
\end{itemize}

As it was difficult to instruct and act out post-stroke survivors' acted-out non-smooth motions, we excluded to act-out \textit{`Smoothness'} component during the study. A researcher, who facilitated the evaluation experiment, only manually indicated a starting cue to start performing an exercise trial. All other functionalities of our system (Figure \ref{fig:setup}) were operated autonomously during the study. The protocols of this user study were reviewed and approved by the Institutional Review Board. 

%% file: contents/results-system.tex
\def\@fnsymbol#1{\ensuremath{\ifcase#1\or *\or \dagger\or \ddagger\or
   \mathsection\or \mathparagraph\or \|\or **\or \dagger\dagger
   \or \ddagger\ddagger \else\@ctrerr\fi}}

\newcommand{\ssymbol}[1]{^{\@fnsymbol{#1}}}
\makeatother

\section{Results}
\subsection{In-Lab System Performance}
\begin{figure}[htp!]
\centering
\resizebox{0.9\textwidth}{!}{%
\centering
  \includegraphics[width=1.0\textwidth]{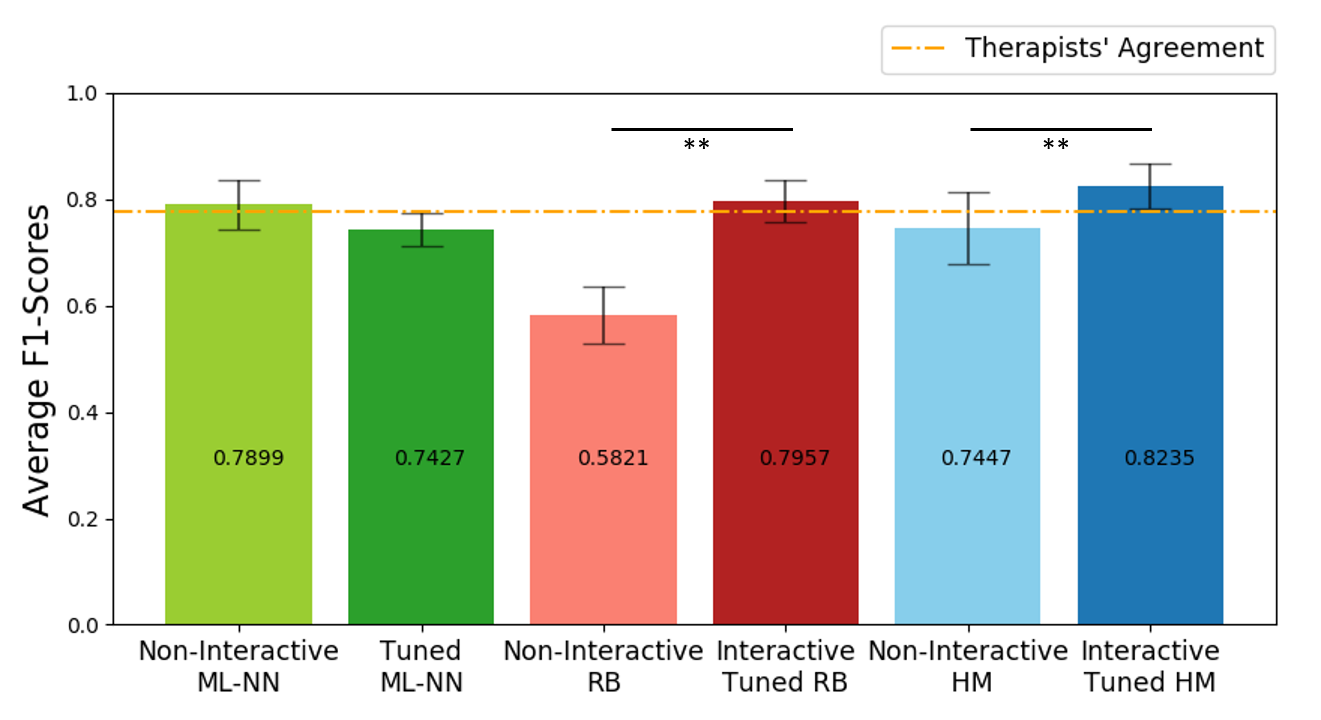}
}
\caption[Comparison of model performance without/with tuning with user data]{Comparison of model performance without/with tuning with user data: Both the rule-based (RB) model and the hybrid model (HM) significantly improved their performance to replicate the therapist's assessment while tuning with patient's unaffected motions. The RB model significantly improved its performance by 37\% from 0.5821 to 0.7957 average F1-score ($p < 0.01$ using paired t-tests) and the HM improved its performance by 11\% from 0.7447 to 0.8235 average F1-score over three exercises ($p < 0.01$ using paired t-tests). {In contrast to the RB and HM models, the Tuned ML-NN performed worse than the ML-NN after tuning with the patient's unaffected motions.}}~\label{fig:results-study-tuning-data-comp}
\end{figure}

Figure \ref{fig:results-study-tuning-data-comp} summarizes the performances of models, which measure an agreement with ground truth labels by computing average F1-scores on performance components of three exercises.
For machine learning (ML) models, we explored a non-interactive, feed-forward neural network (ML-NN), building upon the results from \cite{lee2020towards}. In addition, we presented the results of a tuned, feed-forward neural network (Tuned ML-NN). The parameters of ML-NN models (i.e. hidden layers/units and learning rates of feed-forward neural networks) that achieved the best F1-score during leave-one-subject-out (LOSO) cross-validation are summarized in Appendix. Table \ref{tab:params-ml}. 

In addition, we present the performance of the initial,  non-interactive rule-based model (Non-interactive RB) from the interviews with therapists and that of the interact fine-tuned rule-based model (RB-tuned) after leveraging the held-out user's unaffected motions to tune threshold values for a personalized assessment. The parameters of rule-based models (i.e. the range of the threshold values with $2\sigma$ or $3\sigma$) are selected to achieve the best F1-score during validation: $3\sigma$ is utilized over three performance components of three exercises except for the \textit{`ROM'} and \textit{`Smoothness'} of both Exercise 1 and 2. 

For hybrid models (HMs), we describe the performance of the initial, non-interactive hybrid model (Non-Interactive HM) that integrates the feed-forward neural network (ML-NN) with the non-interactive rule-based model (Non-Interactive RB) and that of the interactive, tuned hybrid model (HM-tuned) that combines the ML-NN with the interactive, tuned rule-based model (RB-Tuned). 

For machine learning (ML) models, Neural Networks (ML-NN) achieve a good agreement level with ground truth annotations (i.e. 0.7899 average F1-score over all exercises), which is equally good with experts' agreement. 
However, the initial, non-interactive rule-based model (Non-Interactive RB) achieves the lowest performance: 0.5827 average F1-score over all exercises. According to the further analysis of the non-interactive, rule-based model, we found that such low performance occurred, because elicited rules from therapists are generic and not tuned for individuals with different physical conditions \cite{lee2020towards}. For instance, one rule of assessing the \textit{`Compensation'} performance component is to check whether the x-coordinate of a shoulder joint is located more than the $15\%$ of the initial position. We found that even if affected motions of some patients were annotated as normal and did not involve compensated shoulder movements, the shoulder joint of those motions was located around $20\%$ of the initial positions, which was greater than a generic threshold value and was misclassified as compensated motions. This indicates the importance of generating personalized rules for patients with various physical characteristics and functional abilities.

The initial, non-interactive hybrid model (Non-Interactive HM) achieved a 0.7447 average F1-score over all exercises. As the initial, non-interactive rule-based model (Non-Interactive RB) had limited performance, the non-interactive HM that integrates the ML model with Neural Networks (ML-NN) and the non-Interactive RB led to slightly lower performance than that of the ML-NN (i.e.  0.7899 average F1-score over all exercises). However, the non-interactive HM still achieved comparable performance to the experts' agreement. 

To evaluate the feasibility of tuning a model for personalized assessment, we updated the threshold values of a rule-based model with held-out patient's unaffected motions (as described in Section \ref{sect:method-rb}) and implemented the interactive, tuned rule-based model (RB-Tuned) and interactive, tuned hybrid model (HM-Tune) that integrates the ML-NN model with the interactive, RB-Tuned model. In addition, we implemented the tuned neural network model (Tuned ML-NN) that fine-tunes a neural network model (ML-NN) with the patient's unaffected motions using the common transfer learning technique \cite{zhuang2020comprehensive}. We then compared the performance of the Tuned ML-NN with that of the interactive, HM-Tuned to evaluate the value of our interactive HM for personalized assessments.

Both RB-Tuned and HM-Tuned models significantly improved their performance to replicate the therapist's assessment ($p < 0.01$ using the paired t-tests over three performance components of three exercises). Specifically, the RB model significantly improved its performance around $37\%$ from 0.5821 to 0.7957 average F1-scores over all exercises ($p < 0.01$). In addition, the hybrid model (HM) also significantly improved its performance around $11\%$ from 0.7447 to 0.8235 average F1-scores over all exercises ($p < 0.01$) and outperformed other approaches. The performance of the tuned hybrid model (HM-tuned) was better than those of the machine learning model with Neural Networks (ML-NN), the Tuned ML-NN, and the RB-Tuned (i.e. $4\%$, $10\%$, and $3\%$ improvement respectively without statistical significance). Unlike our RB-Tuned and HM-Tuned models that improved their performance, the Tuned ML-NN performed worse 5\% from 0.7899 to 0.7427 average F1-scores after tuning with the patient's unaffected motions.

\begin{figure}[h!]
\centering
\resizebox{0.8\columnwidth}{!}{%
  \includegraphics[width=1.0\columnwidth]{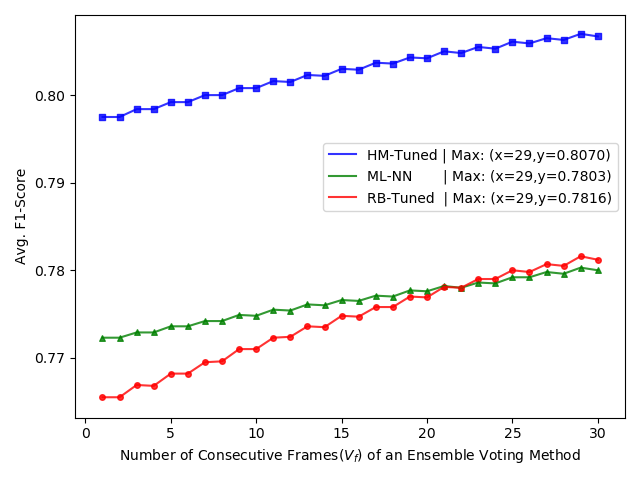}
  }
  \caption{Performance of frame-level assessment with different numbers of consecutive frames ($V_f$) using the tuned rule-based model (RB-Tuned), machine learning model with neural networks (ML-NN), tuned hybrid model (HM-Tuned)}
  \label{fig:results-voting-comp}
\end{figure}

To analyze the effect of our ensemble voting method for frame-level assessment, we utilized the ML-NN, RB-Tuned, and HM-Tuned models and plotted their average performance of detecting frame-level compensation on the head, spine, shoulder joints over three exercises with various numbers of consecutive frames ($V_f$ = {1, ..., 30}). In Figure \ref{fig:results-voting-comp}, all three models (i.e. ML-NN, RB-Tuned, HM-Tuned) improved their performance while leveraging prediction from multiple frames and achieved their best performance with $V_f=29$. When we compared the performance of a model without and with an ensemble voting method ($V_f = 1$ and $V_f = 29$), the ML-NN model improved its performance from 0.7723 ($V_f=1$) to 0.7803 ($V_f=29$) average F1-score ($p < 0.01$ using the paired t-tests over three compensations of three exercises); the RB-Tuned model improved its performance from 0.7655 ($V_f=1$) to 0.7816 ($V_f=29$) average F1-score ($p < 0.01$); the HM-Tuned model improved its performance from 0.7975 ($V_f=1$) to 0.8070 ($V_f=29$) average F1-score ($p < 0.01$).

\subsection{Real-World System Performance}\label{sect:results-therapists}

\begin{figure}[tp!]
\centering
\resizebox{0.75\textwidth}{!}{%
\centering
  \includegraphics[width=1.0\textwidth]{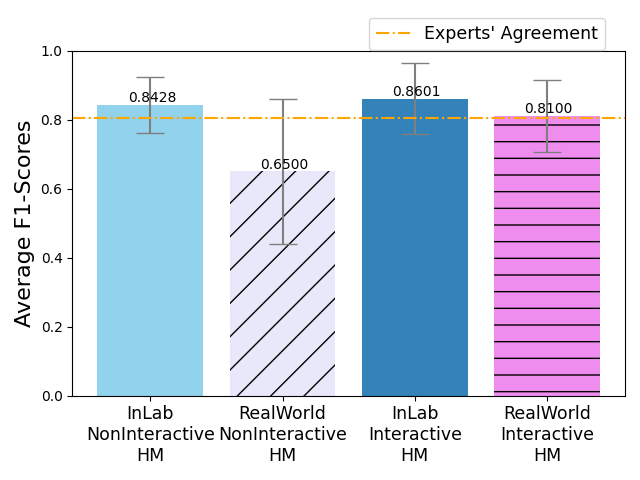}
}
\caption[Comparison of in-lab and real-world evaluations]{Comparison of performance of non-interactive and interactive hybrid models for quantitative rehabilitation assessment during in-lab and real-world studies: Although our interactive hybrid model achieved slightly lower performance during a real-world study, our interactive model can still adapt to a new participant and achieve comparable performance to experts' agreement unlike non-interactive models. Compared to the non-interactive HM model, our interactive HM model had much lower performance degradation from the in-lab study to the real-world study.}~\label{fig:results-performance-realworld}
\end{figure}

Figure \ref{fig:results-performance-realworld} summarizes the performance of non-interactive and interactive hybrid models during the in-lab and real-world studies. Our results showed that non-interactive models of the real-world study led to lower average performance compared to the models of the in-lab study along with the performance degradation of 22\% from 0.84 F1-score to 0.65 F1-score. Also, interactive models of the real-world study led to lower average performance than the models of the in-lab study. However, our interactive models led to a performance degradation of 5\% from 0.86 F1-score to 0.81 F1-score, which is less than that of non-interactive models. Our system could still adapt to new participants with diverse physical characteristics and achieved performance that is comparable with experts' agreement.

\begin{figure}[htp!]
\centering
\resizebox{0.95\textwidth}{!}{%
\centering
  \includegraphics[width=1.0\textwidth]{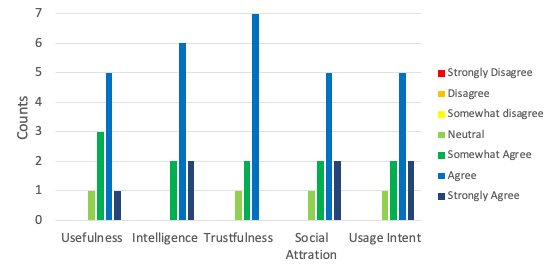}
}
\caption{Usability survey responses from participants during the real-world evaluation study: most participants were positive to use our system which is considered as socially attractive and intelligent by providing useful and trustful information.}~\label{fig:results-usability-realworld}
\end{figure}

In addition, Figure \ref{fig:results-usability-realworld} describes the histogram of usability responses from participants in the real-world study. Overall, participants in the real-world study expressed positive opinions on our interactive robotic exercise coach. They appreciated our system that is  \textit{``useful to observe my [their] body alignment and positions''} (P5) during exercises and provide feedback on \textit{``how well I [a participant] did it [an exercise]''} (P6). Even if our system provides incorrect assessment, participants considered that feature-specific feedback from our system assisted them to better determine whether to trust the feedback of a system: \textit{``I find the system trustful because the feedback made me clear how I should improve in a certain way''} (P9). Participants also enjoyed having an exercise with a robot, but one participant considered that it would be better if the robot can have \textit{``a more friendly face or appearance''}(P3). Overall, participants were positive to use or recommend our system as an exercise coach even with some limitations.

%% file: contents/discussion.tex
\section{Discussion}
In this work, we study and discuss how a social robot exercise coaching system can be designed and developed to generate personalized corrective feedback along with the in-lab and real-world evaluation studies.

For generating personalized corrective feedback, we compared various existing approaches with our proposed hybrid model and evaluated the effect of an ensemble voting method for real-time, frame-level assessment \cite{lee2020towards}. Among various approaches, the machine learning model with Neural Networks (ML-NN), the tuned rule-based model (RB-tuned), and the initial, non-interactive and the interactive, tuned hybrid models (Non-Interactive HM and Interactive, Tuned HM) have equally good performance with expert's agreement from the paired t-tests over three performance components of three exercises. In addition, all models with an ensemble voting method can leverage predictions from multiple consecutive frames to improve their frame-level assessment and inform a user when an erroneous motion has occurred. 

As a rule-based (RB) model does not require the data collection process, an RB model could be considered as a natural starting point to develop a social robot exercise coaching system that can assess the quality of motion and generate corrective feedback on patient's exercises \cite{mataric2007socially,lee2020towards}. However, an RB model with generic threshold values (e.g. RB-Init and \cite{fasola2013socially,gorer2017autonomous,lee2020study,guneysu2017socially,tanguy2016computational}) does not perform well to evaluate exercises of patients with various physical conditions. Thus, it is important to have an interactive approach that can tune an RB model with individual's held-out unaffected motions to derive personalized threshold values for assessment and corrective feedback. 

When a social robot exercise coaching system is deployed and annotated data is collected, a machine learning (ML) model (e.g. neural networks) can be trained to extract new insights for assessing exercises from data. However, we do not recommend simply replacing a rule-based (RB) model with an ML model using a complex algorithm that operates as a black box model. For instance, given a patient's affected motion that is incorrectly performed with compensation, an ML model with Neural Networks can just notify whether the compensation has occurred or not without any explanations on the outputs of the model \cite{rudin2019we}. In contrast, our interactive hybrid model can predict assessment with improved performance, but also identify which feature has been violated with a rule-based model: the violation on the head in the z-axis and the shoulder in the y-axis for Figure \ref{fig:sample-compensation-abnormal-e1-p14}. Such feature-level analysis can be realized in the following personalized corrective feedback: \textit{``Keep your head straight and do not raise your shoulder''} \cite{lee2020towards,lee2022enabling}. We found that participants in our real-world study appreciated the potential of our system to make them have trustful interactions with it. Thus, after data collection, a hybrid model is recommended to accommodate new generic insights from data and support a transparent and personalized interaction between a robot and a user. 

When it comes to the evaluation of the system performance, we found that our in-lab study through leave-one-subject-out (LOSO) cross-validation has a slightly over-promising performance than our real-world study. However, the performance difference is not statistically significant using the t-test. Thus, we considered that the LOSO cross-validation has the potential to provide the estimated system performance in practice, which still needs to be carefully analyzed further \cite{rao2008dangers}.

During our real-world study, we found that our interactive robotic exercise coach has the potential to adapt to a new user and automatically monitor participants' exercises and provide personalized corrective feedback. However, our system implementation still requires manual input from a researcher to indicate the starting time of the user's motion. For creating a fully autonomous system, the exploration of techniques for motion segmentation \cite{lin2013online} is necessary. In addition, we have only conducted the pilot evaluation with healthy participants, who acted out post-stroke survivor's motions. In-person user studies with post-stroke survivors are required to better understand the feasibility of our system in practice. As post-stroke survivors might perform incorrect motions that might exacerbate their conditions, it is also important to explore a way to adapt a rehabilitation session and program \cite{lee2022enabling} beyond personalized feedback that has been studied in this work. 

%% file: contents/conclusion.tex
\section{Conclusion}
In this paper, we contributed to the designs, development, and evaluation of an interactive approach with an ensemble voting method for a social robot exercise coach system in the context of physical stroke rehabilitation therapy. This system integrates a machine learning model with an interactive and interpretable rule-based model and tunes with patient's data for real-time, personalized corrective feedback on patient's exercises. Through in-lab and real-world experiments, this work shows that our interactive hybrid model can adapt to a new user and achieve better performance to replicate an expert's assessment and feedback on unobserved data of new users, but also support transparent and personalized interaction of a robotic exercise coaching system. In addition, this work discusses the potential benefits and limitations of our system to support post-stroke survivor's rehabilitation sessions.

%% file: contents/appendix.tex
\appendix
\section{Appendix}

\begin{table}[h]
\centering
\caption{Parameters of Machine Learning Models}
\resizebox{0.84\columnwidth}{!}{%
\begin{tabular}{clll}\toprule
 & \multicolumn{3}{c}{Hidden Layers and Units / Learning Rate} \\ \midrule
 & \multicolumn{1}{c}{ROM} & \multicolumn{1}{c}{Smoothness} & \multicolumn{1}{c}{Comp} \\ \midrule
E1 & \begin{tabular}[c]{@{}l@{}}- NN: (256, 256, 256) / 0.005\end{tabular} & \begin{tabular}[c]{@{}l@{}}- NN: (16) / 0.0001\end{tabular} & \begin{tabular}[c]{@{}l@{}}- NN: (512, 512, 512)  / 0.005\end{tabular} \\ \midrule
E2 & \begin{tabular}[c]{@{}l@{}}- NN: (32, 32, 32) / 0.01\end{tabular} & \begin{tabular}[c]{@{}l@{}}- NN: (32) / 0.0001\end{tabular} & \begin{tabular}[c]{@{}l@{}}- NN: (256, 256) / 0.0001\end{tabular} \\ \midrule
E3 & \begin{tabular}[c]{@{}l@{}}- NN: (16) / 0.005\end{tabular} & \begin{tabular}[c]{@{}l@{}}- NN: (128) / 0.0001\end{tabular} & \begin{tabular}[c]{@{}l@{}}- NN: (256, 256, 256) / 0.1\end{tabular}\\ \bottomrule
\end{tabular}%
}
\label{tab:params-ml}
\end{table}

\begin{table}[h]
\caption{Performances (avg. $\pm$ std. of F1-scores) of machine learning (ML) models, rule-based (RB) models, hybrid models (HMs), and experts' agreement. $\ssymbol{3}$ indicates HM-Tuned performs statistically better than the compared method (pairwise t-tests at 99\% significance level).}
\label{tab:results-avg}
\resizebox{\columnwidth}{!}{%
\begin{tabular}{ccccc} \toprule
\multirow{1}{*}{Algorithm} & \multirow{1}{*}{Exercise 1} & \multirow{1}{*}{Exercise 2} & \multirow{1}{*}{Exercise 3} & \multirow{1}{*}{Overall} \\ \midrule
ML-NN &  {0.8428 $\pm$ 0.0809} & {0.7549 $\pm$ 0.1026} & {0.7720 $\pm$	0.0433} & {0.7899 $\pm$ 0.0466} \\
Tuned ML-NN &  {0.7707 $\pm$ 0.1093} & {0.7105 $\pm$ 0.1308} & {0.7470 $\pm$ 0.0381} & {0.7427 $\pm$ 0.0303}  \\
RB-Init $\ssymbol{3}$  & 0.6148 $\pm$ 0.2086 & 0.6707 $\pm$ 0.1758 & 0.4626 $\pm$ 0.2102 & 0.5827 $\pm$ 0.0541 \\
RB-Tuned & 0.8317 $\pm$ 0.0784 & 0.8009 $\pm$ 0.1238 & 0.7543 $\pm$ 0.0248 & 0.7957 $\pm$ 0.0390 \\  \midrule  \midrule
HM-Init $\ssymbol{3}$ &  0.8069 $\pm$ 0.0946 & 0.7060 $\pm$ 0.1318 & 0.7212 $\pm$ 0.0851 & 0.7447 $\pm$ 0.0679 \\
HM-Tuned & {0.8601 $\pm$ 0.1030} & {0.7769 $\pm$ 0.1317} & {0.8334 $\pm$ 0.1142} & \textbf{0.8235 $\pm$ 0.0425} \\  \midrule \midrule
\begin{tabular}[c]{@{}c@{}}Experts'\\Agreement\end{tabular} & {0.7908 $\pm$ 0.2146} & {0.8222 $\pm$ 0.1534} & {0.7196 $\pm$ 0.1754} & {0.7775 $\pm$ 0.0526} \\ \bottomrule 
\end{tabular}%
}
\end{table}

\input{tables/list-rules}

%% file: tables/list-rules.tex
\begin{table}[h]
\caption[rules]{List of independent rules to assess the quality of motion from therapists}
\label{tab:method-rules-all}
\resizebox{\textwidth}{!}{%
\begin{tabular}{cl} \toprule
\textbf{\begin{tabular}[c]{@{}c@{}}Performance\\ Components\end{tabular}} &
  \multicolumn{1}{c}{\textbf{Rules}} \\ \midrule
\multirow{3}{*}{\begin{tabular}[c]{@{}c@{}}Range of  Motion\\ (ROM)\end{tabular}} &
  a wrist joint should be located above a spine-shoulder joint near a head joint for exercise 1 \\
 &
  a wrist joint should be located higher than a shoulder joint for exercise 2 \\
 &
  a wrist joint should be located further than hip near a knee for exercise 3 \\ \midrule
\multirow{3}{*}{Smoothness} &
  \begin{tabular}[c]{@{}l@{}}a wrist joint should be smoothly coordinated in the x-axis during 80\% of the motion\\          (zero-crossing ratio of a wrist acceleration in the x-axis is within 20\%)\end{tabular} \\
 &
  \begin{tabular}[c]{@{}l@{}}a wrist joint should be smoothly coordinated in the y-axis during 80\% of the motion\\          (zero-crossing ratio of a wrist acceleration in the y-axis is within 20\%)\end{tabular} \\
 &
  \begin{tabular}[c]{@{}l@{}}a wrist joint should be smoothly coordinated in the z-axis during 80\% of the motion\\          (zero-crossing ratio of a wrist acceleration in the z-axis is within 20\%)\end{tabular} \\ \midrule
\multirow{9}{*}{Compensation} &
  a head joint should not be located more/less than 15\% of an initial head position in the x-axis \\
 &
  a head joint should not be located above/below 15\% of an initial head position in the y-axis \\
 &
  a head joint should not be located more/less than 15\% of an initial head position in the z-axis \\
 &
  a spine joint should not be located more/less than 15\% of an initial spine position in the x-axis \\
 &
  a spine joint should not be located above/below 15\% of an initial spine position in the y-axis \\
 &
  a spine joint should not be located more/less than 15\% of an initial spine position in the z-axis \\
 &
  a shoulder joint should not be located more/less than 15\% of an initial shoulder position in the x-axis \\
 &
  a shoulder joint should not be located above/below 15\% of an initial shoulder position in the y-axis \\
 &
  a shoulder joint should not be located more/less than 15\% of an initial shoulder position in the z-axis \\ \bottomrule
\end{tabular}%
}
\end{table}